\documentclass{article}

\usepackage{microtype}
\usepackage{graphicx}
\usepackage{subfigure}
\usepackage{booktabs} % for professional tables
\usepackage{amsmath}
\usepackage{multirow}
\usepackage{amssymb}

\newcommand{\pb}[1]{\textcolor{blue}{[pb: #1]}}

\newcommand{\mT}[0]{\mathcal{T}}

\usepackage{hyperref}

\usepackage[accepted]{icml2019}

\icmltitlerunning{NeuralSampler}

\begin{document}

\twocolumn[
\icmltitle{NeuralSampler: Euclidean Point Cloud Auto-Encoder and Sampler}

\begin{icmlauthorlist}
\icmlauthor{Edoardo Remelli}{epfl}
\icmlauthor{Pierre Baque}{epfl}
\icmlauthor{Pascal Fua}{epfl}
\end{icmlauthorlist}

\icmlaffiliation{epfl}{CVLab, EPFL, Lausanne, Switzerland}

\icmlcorrespondingauthor{Edoardo Remelli}{edoardo.remelli@epfl.ch}

\icmlkeywords{Machine Learning, ICML}

\vskip 0.3in
]



\printAffiliationsAndNotice{}  % leave blank if no need to mention equal contribution

\begin{abstract}

Most algorithms that rely on deep learning-based approaches to generate 3D point sets can only produce clouds containing fixed number of points. Furthermore, they typically require large networks parameterized by many weights, which makes them hard to train. 

In this paper, we propose an auto-encoder architecture that can both encode and decode clouds of arbitrary size and demonstrate its effectiveness at upsampling sparse point clouds. Interestingly, we can do so using less than half as many parameters as state-of-the-art architectures while still delivering better performance.  We will make our code base fully available.
\end{abstract}

% !TEX root = top.tex
% !TEX spellcheck = en-US

\section{Introduction}

Point clouds are becoming increasingly popular as a compact and expressive way to represent 3D surfaces because they can capture high frequency geometric details without requiring much memory. As in many other areas of computer science, there are now many deep learning approaches that can process and generate such data structure efficiently. However, while recent architectures such as PointNet~ \cite{Qi17a} can handle inputs with non-fixed dimension, most cloud generative models~\cite{Achlioptas18, Yang18a, Zamorski18} produce outputs with a fixed number of elements. This precludes changing the topology at run-time and increasing the level of details where needed and only there. In theory, this could be addressed by increasing the dimension of the output. This, however, would make the already slow and computationally intensive training of deep networks even more so to the point of impracticality.

As an important step towards breaking this bottleneck, we propose a novel convolutional auto-encoder architecture for 3D cloud representation learning, which we will refer to as \textit{NeuralSampler}, that can both encode and decode clouds of arbitrary size and demonstrate its effectiveness at upsampling sparse point clouds. Unlike earlier approaches, our methods decouples \textit{shape generation} from \textit{surface sampling}, learning representations that can produce clouds made of arbitrarily many points.

\begin{figure}[t]
	\begin{center}
		\vskip -0.2in
		{\includegraphics[width=0.9\columnwidth]{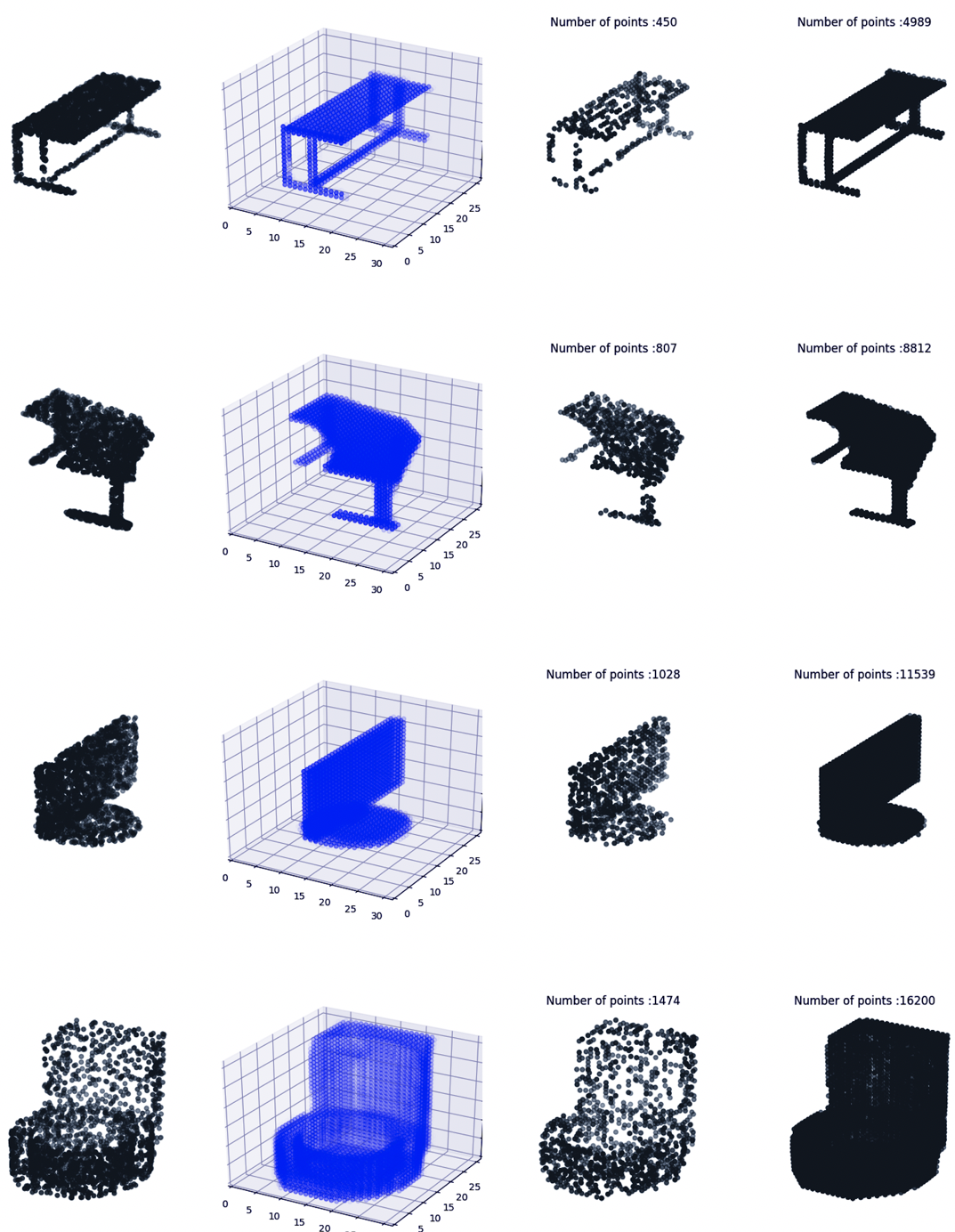}}\\
		(a) \hspace{1.6cm}(b)  \hspace{1.6cm}(c)  \hspace{1.6cm}(d)
		\vskip -0.1in
		\caption{\textbf{Upsampling objects from ModelNet dataset}~\cite{Wu16b}. (a) Input 3D points. (b) Occupancy maps. (c) 3D points generated after one single forward pass. (d)  3D points generated after 10 forward passes though the sampling layer.  }
		\vskip -0.2in
		\label{fig:teaser}
	\end{center}
	
\end{figure}

More specifically and as shown in Figure \ref{fig:teaser}, we first define a voxel grid containing the point clouds to be modeled in order to capture global structure and reduce the total number of parameters in our auto-encoder.
We then train an auto-encoding architecture to yield an estimate of probability of a sample point to fall within each voxel -visualized in column (b) of Figure \ref{fig:teaser} - which we will refer to as an occupancy map, and a feature vector describing its local properties. The latter is then fed to a sampling layer that produces a non-deterministic set of 3D offsets describing the position of the points with respect to the center of the grid cell.  The cardinality of the point set output by our decoder depends on the estimated surface area. Note that, once we generated a shape, we can sample from it repeatedly, allowing us to  generate 3D point clouds with arbitrary spatial resolution, as highlighted in column (d) of Figure \ref{fig:teaser}. 

Our contribution is therefore an approach to learning a 3D point cloud representation that can produce high resolution point clouds while still harnessing the power of deep learning. This enables NeuralSampler to achieve good precision even when working with coarse underlying voxel grids and to outperform state-of-the-art super-sampling architectures on a standard representation learning 3D point cloud classification benchmark~\cite{Achlioptas18}. Furthermore, we can do so using less than half as many parameters of any previous method. 

The rest of this paper is organized as follows. Section \ref{sec:related} reviews related prior work. Section \ref{sec:method} present our fully convolutional approach to learning a 3D point cloud representation. In Section \ref{sec:exp} we evaluate our model both qualitatively and quantitatively over standard benchmarks. Lastly in Section \ref{sec:conclusions} we draw conclusions and propose further applications of our method. To foster progress of research in this area, we will make our code base fully available on the project web page.

% !TEX root = top.tex
% !TEX spellcheck = en-US

\section{Related Work}
\label{sec:related}

In computer graphics, computer vision, computer-aided design, and computer-aided manufacturing, surfaces have been represented in terms of volumetric primitives, wireframes, triangulated meshes, implicit surfaces, and 3D point clouds. Recently, in the Deep Learning community, there has been renewed interest in the latter. They make it possible to work with detailed representations while requiring very little memory, but at the cost of giving up the regular grid structure that 
 standard deep learning frameworks, such as Convolutional Neural Networks (CNNs), depend on.

The PointNet architecture~\cite{Qi17a} overcomes this difficulty. Its ability to handle unstructured sets of 3D points has sparked the development of new architectures that can operate on point clouds without {\it any} structure while still taking into account neighborhood relationships~\cite{Qi17b,Wang18b,Yu18a,Achlioptas18}. They have been shown to be effective for tasks such as point cloud segmentation and classification. However, comparatively little work has focused on devising deep architectures to generate 3D point clouds and we review the existing approaches below. 

One of the first approaches~\cite{Nash17}  to generating clouds relies on a variational auto encoder that outputs for each point 3D coordinates, a normal estimate, and an object part probability. It encodes and decodes this representation into and from a latent vector using fully connected layers. This approach is refined in~\cite{Achlioptas18} by using the PointNet architecture~\cite{Qi17a} to encode the latent vectors.  In~\cite{Zamorski18}, the focus is on improving the accuracy of generated samples by relying on an adversarial paradigm. All these approaches use  fully connected layers to regress fixed size point clouds from the latent vectors, which requires millions of parameters to generate a relatively small and fixed number of points and are therefore hard to train. This makes them computationally expensive and overfitting prone. Exceptions to this are the approaches of~\cite{Yang18a} and~\cite{Yu18a}. The first introduces a folding decoder designed for point cloud generation and reduces by one order of magnitude the required number of parameters. However, the decoder processes each point independently, thus ignoring neighborhood relationships, and still outputs a fixed number of points.
The second uses an upsampling scheme based on feature expansion, and grouped PointNet convolutions, all of which reduces the total number of parameters. However, because it does not use a latent representation, the number of parameters remains large. 

Closest in spirit to our approach, is that of~\cite{Li18e} that relies on Generative Adversarial Networks to learn to generate point cloud distributions using a hierarchical sampling process. However, it differs from ours by its use of a fully connected layers to generate only one surface point per forward pass, which probably explains why their reported performance on the representation learning benchmarks is below the current state-of-the-art.

%\er{the paper is extremely hard to follow and does not give any information on the kind of architecture they employ, nor really details how their sampling process works...}

% !TEX root = top.tex
% !TEX spellcheck = en-US

\section{Method}
\label{sec:method}

\begin{figure}[h!]
	\begin{center}
		\vskip -0.1in
		{\includegraphics[width=\columnwidth]{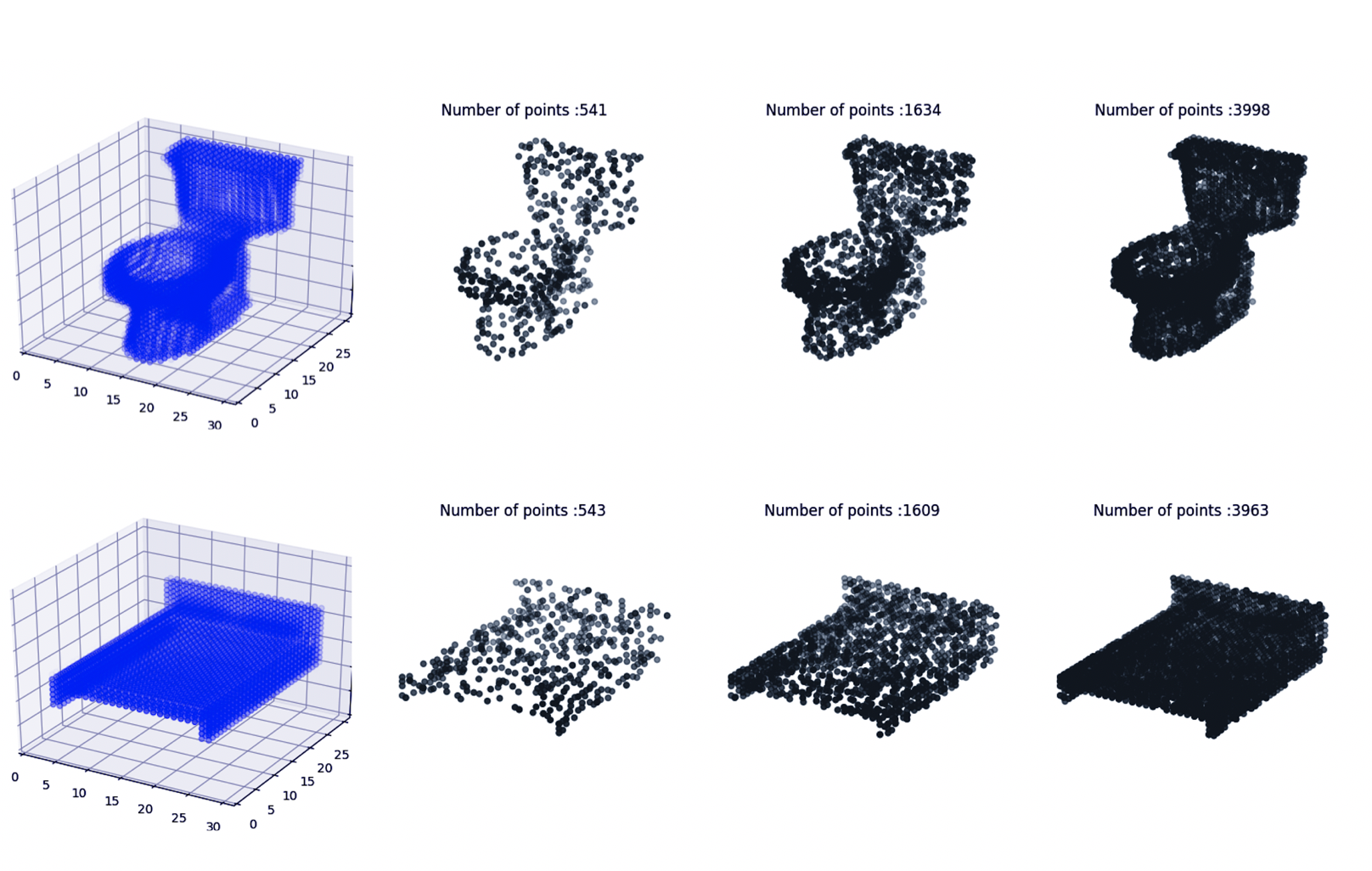}}
		\vskip -0.1in
		\caption{\textbf{Sampling at increasing resolution.} From left to right, occupancy map, point clouds generated by 1, 3, 6 forward passes. Note that our method effectively produces different surface samples for each forward pass. }
		\label{fig:sampling}
	\end{center}
	\vskip -0.2in
\end{figure}

In this section, we present our fully convolutional approach to learning a 3D point cloud representation that can be used to generate point clouds with an arbitrary spatial resolution. Our goal is to train an encoder-decoder architecture that given a cloud $Y_{\text{in}} \in \mathbb R ^{P \times 3}$ obtained by sampling a 3D shape $\mathcal S$ represented by a surface mesh,  generates a new point cloud $Y_{\text{out}}  \in \mathbb R ^{Q \times 3}$ that is also a sampling of surface $\mathcal S$. The difference between $Y_{\text{in}}$ and $Y_{\text{out}}$ is that the first contains $P$ points and the second $Q$, where $P$ and $Q$ are arbitrary and can vary from shape to shape. 

 Given a input surface sampling $Y_{\text{in}}$ for each shape, we compute occupancy maps such as those shown in the first column of Figure ~\ref{fig:sampling} along with local feature vectors. These are fed to sampling layers that generate 3D points, or not, depending on the occupancy value. Figure ~\ref{fig:sampling} depicts this sampling process. The rightmost columns depict the results after repeating this operation 1, 3 and 6 times. 
 
 We now introduce our approach to parameterizing point clouds on Euclidean grids and then describe our convolutional encoding-decoding architecture. 

\begin{figure*}[ht]
	\vskip -0.12in
	\begin{center}
		{\includegraphics[width=\textwidth]{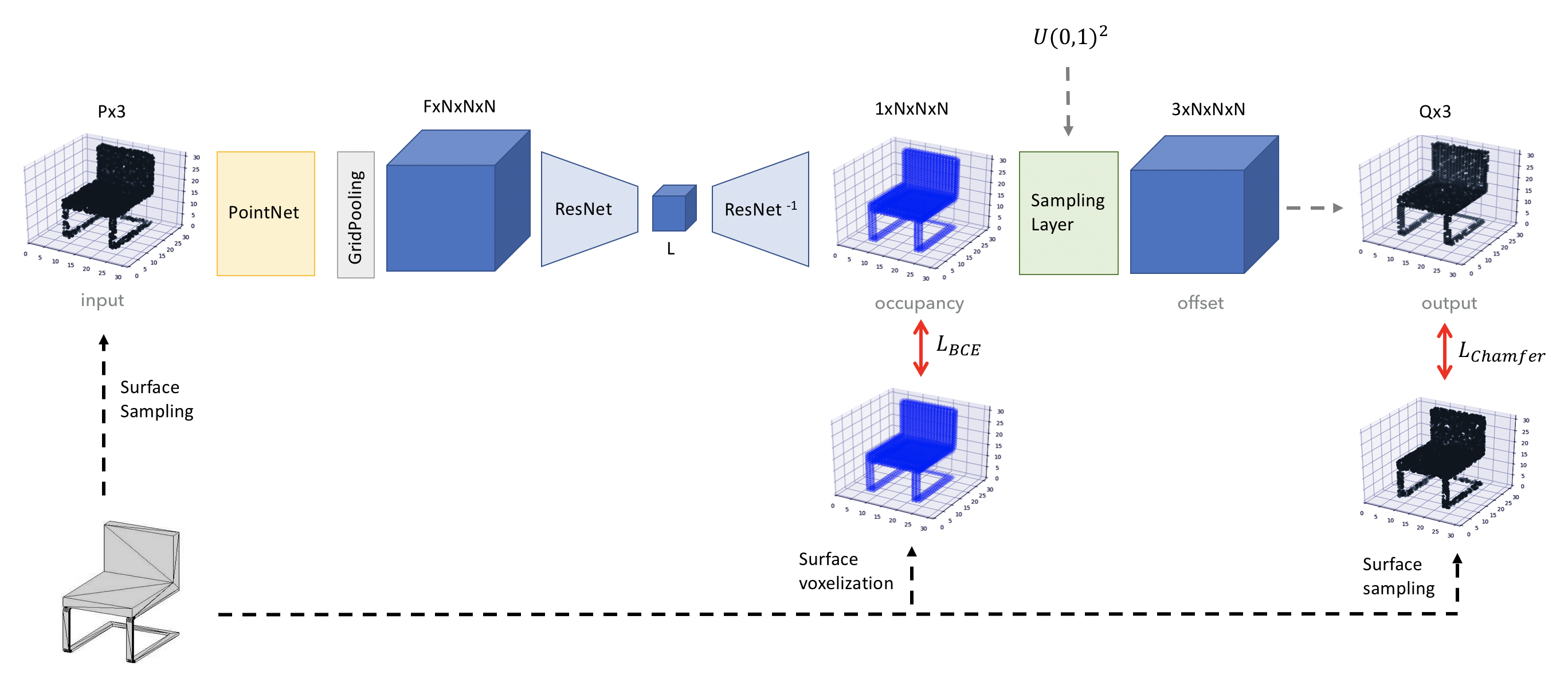}}
		\vspace{-8mm}
		\caption{\textbf{NeuralSampler Architecture.} Given an input point cloud, we embed it into a coarse 3D grid, and map it to a latent vector representation by combining PointNet layers and a shallow ResNet architecture. We then map the latent vector back into the grid whose voxels contain an estimate of the probability of a sample to fall within the cell, which we refer to as the occupancy, and a local feature vector. Finally, in non-empty voxels, we use the feature vector in conjunction with 2 randomly chosen  coordinates to produce a 3D surface sample. For each forward pass, the probability that a point will actually be generated depends on the occupancy value.}
		\vskip -0.2in
		\label{fig:architecture}
	\end{center}
\end{figure*}

%This is a desirable to avoid under- or over-sampling when working with datasets featuring large scale variations or when only some samples contains high frequency details. 

%This is a desirable property to avoid under/over-sampling issues when we target representation learning for datasets containing large scale variations, or when dealing with datasets where only some samples contains high frequency details. Moreover, as we will illustrate, this property can be used to extend our approach to point cloud super-sampling applications, as we will illustrate in Section \ref{sec:exp}.

\subsection{Using Grids to Parameterize Point Clouds}
\label{sec:param}

The 3D points that form the cloud represent a potentially irregular sampling of a 3D surface and we need an object-centered representation that can handle surfaces of arbitrary complexity and topology. In the computer vision and computer graphics literature, this has been addressed by describing the surfaces in terms of local 3D patches. They can either be planar and located in a voxel grid ~\cite{Szeliski92a,Fua97a} or more free form ~\cite{Varol12b}. More recently, a 3D Euclidean grid has been used in conjunction with a deep marching cube algorithm ~\cite{Liao18a}. 

Similarly, we embed our 3D point cloud into a voxel grid and represent individual 3D points in terms of a 3D offset with respect to the voxel centers. 

Let us assume, without loss of generality, that each voxel has size $1$, and let us introduce scalar \textit{surface occupancy} field $O \in [0,1]^{1 \times N \times N \times N}$ and a vectorial \textit{offset} field $\Delta \in \mathbb[-1/2,1/2]^{3 \times N \times N \times N}$.

We will now explain how we can first define a point cloud \textit{distribution} $T$ implicitly via surface occupancy field $O$, and then how, combining topological information with offset $\Delta$, we can define a point cloud distribution $X=X(T(O),\Delta)$.

Let us define $t_n \in \{0,1\}$ as the random variable indicating if $n$-th voxel contains a point or not. We parameterize $t_n$ as a Bernoulli distribution with parameter $o_n$, i.e. 
\begin{align}
t_n \sim p(t_n) = o_n ^ {t_n} (1 - o_n) ^ {1 - t_n}.
\end{align}
Assuming each voxel indicator to be independent, we can compute the probability distribution of the random variable $T$, indicating if our point cloud has a given occupancy $T^*$ over the full voxel grid, as
\begin{align}
p( T = T^*) = \prod_{n=1}^{N\times N \times N} p (t_n = T_{n}^*).
\end{align}

Note that, from elementary probability theory, $T$ also follows a Bernoulli distribution.
The scalar occupancy field O can, therefore, be used to span the set $\mT$ of all possible topologies over a $\{ N \times N \times N \}$ grid. Note that this set has cardinality $|\mT| = 2^{N \times N \times N}$. 

Once the point could topological distribution has been captured via the surface occupancy, we can use the offset tensor field to parameterize each individual $n$-th point position within the $n$-th voxel as
\begin{align}
x_n = \bar{x}_n + \Delta_n,
\end{align}
where $\bar{x}_n$ denotes the position of $n$-th cell center.

To recap, we have shown how $T$ and $\Delta$ can be used jointly to define a distribution over point clouds. Note that the number of elements in a point cloud defined with such parameterization depends on both the resolution of the grid and the area of the ground truth surface. In the upcoming section we will illustrate how, however, surface sampling allows our formulation to handle high resolution point clouds with relatively coarse Euclidean grids.

\comment{
{Is $\Delta$ a random variable as well?} $X(T,\Delta)$, which, from elementary probability theory, under the assumption that $P(T_n=0) \neq 0.5 \, \forall n \in \{N \times N \times N \} $, follows a unimodal distribution. We will see that this observation, in practice, is crucial to make computations tractable when optimizing the parameters of our network. \pb{What do you call unimodal distribution in this context? This is not clear}. }

\subsection{Convolutional Encoder Architecture}

We propose to use the encoder architecture introduced in ~\cite{Liao18}, which maps a raw input point cloud $Y$ into a latent code $z = e(Y)$ via a convolutional network. 
Similar to PointNet ~\cite{Qi17a}, we first extract a local feature vector for each point using 1D convolutions in feature space. 
We then aggregate features onto an auxiliary $  N \times N \times N$ 3D grid, by grouping together all point falling within a cell and applying max-pooling within each cell. 
We will refer to this operation as \textit{grid pooling}. 
Once features are aggregated onto a 3D grid, we use a shallow ResNet ~\cite{He16} to extract global structure across grid cells and map to a low dimensional latent embedding $z$. 
More detailed information on feature dimensions and number of layers are provided in the supplementary material.

\subsection{Convolutional Decoder Architecture}
Our approach is motivated by the idea that point clouds are random samplings of geometric surfaces, and handle their reconstruction in a non-deterministic setting, rather than aiming at reproducing exactly the 3D coordinates of the input point set.  

To achieve this, we propose a decoder architecture which casts the task of generating point clouds into a conditional setting: first, it generates a surface descriptor for ground truth shape $\mathcal S$, and, successively, produces random surface offsets $\Delta$ conditioned on the surface descriptor and an additional 2D noise vector $(u,v)$.

 In practice, latent codes are  mapped to the occupancy tensor $O$ introduced in section \ref{sec:param}, together with a feature map $F$ containing, within each voxel, local information about the surface. Once we have produced such surface descriptor, we rely on a \textit{sampling layer}, which takes as input a $F$ and random sampling coordinates $(u,v)$ drawn from a two-dimensional uniform distribution to produce a non deterministic offset value $\Delta$ within each voxel along the local 2D surface manifold. Similar in spirit to the idea introduced in \cite{Yang18a}, this layer maps a 2D grid into a surface manifold. However, in order to handle high frequency details, we do so within each voxel.
 
We can express the likelihood of this joint distribution as
\begin{align}
p(\Delta(u,v), O, F| S) = p(\Delta(u,v)| F, S) \, p(O, F|S).
\end{align}

As both shape descriptors are defined over an Euclidean Grid, our decoder architectures simply reverts all ResNet encoding operations and performs up-sampling using transposed 3D convolutions. We will denote it as ResNet$^{-1}$. 
Concerning the sampling layer, we concatenate the surface descriptor to the sampling coordinates $(u,v)$, and rely on two convolutional layers to map this information into an offset value $\Delta$, as described in section \ref{sec:param}. We refer the reader to the supplementary material for the detailed specification of our decoder architecture.

After having successfully decoupled shape generation from surface sampling, we can, at test time, generate a shape and then sample its surface repeatedly by feeding random sampling coordinates to the sampling layer, allowing us to generate output 3D point clouds with arbitrary spatial resolution.
We will describe in the next section how our training strategy allows us to decouple surface generation and sampling. 

\subsection{Loss Function and Network Training}

\textbf{Reconstruction loss.} Our auto-encoder aims at reconstructing target output point cloud $Y_{\text out}$, given latent embedding $z$ produced from encoding input point cloud $Y_{\text in}$. To ease the notation we will indicate the output point cloud as $Y$, recalling that $Y_{\text in}$ and $Y_{\text out}$ are random samples of the ground truth surface $S$ that can have arbitrary size.

We measure the discrepancy between output point cloud $X(T,\Delta)$ and target point set $Y$ with the so-called Chamfer (pseudo) distance  ~\cite{Fan17a}, defined as 
\begin{align}
d(X(T,\Delta),Y) &= \sum _{x \in X(T,\Delta)} \min_{y \in Y} \| x - y\| \\ &+ \sum _{y \in Y} \min_ {x \in X(T,\Delta)}\| x - y\| \\
&= d(X \mid Y) + d(Y \mid X).
\end{align}
Note that, in our setting, we cannot use the \textit{Earth Mover}'s distance ~\cite{Rubner01} since this metric cannot handle point sets having different cardinality~\cite{Achlioptas18}.

Since the point cloud configuration is non-deterministic, we train our model so that to minimize the expected value of the Chamfer Distance over all possible point clouds distributions, i.e.
\begin{align}
\label{eq:loss}
\min _ \theta \mathbb E_{T, \Delta} \, d(X(T(\theta),\Delta(\theta)),Y),
\end{align}
where $\theta$ are the parameters of our network.

However, if we expand the term above, we see that its optimization is not tractable.

For the first distance term, in fact, we have
\begin{align}
\mathbb E _{T, \Delta} \,d(X| Y) &= \mathbb E _{T, \Delta} \sum _{x \in X(T)} \min_{y \in Y} \| x - y\| \\
&= \sum _{n \in {N \times N\times N}} p(t_n = 1) E _{\Delta} \min_{y \in Y} \| x_n - y\|.
\end{align}

For the second term, it holds
\begin{align}
\mathbb E _{T, \Delta} \,d(Y| X) &= \mathbb E _{T, \Delta} \,\sum _{y \in Y} \min_{x \in X(T)} \| x - y\| \\
&= \sum _{y \in Y} \mathbb E _{T, \Delta} \, \min_{x \in X(T)} \| x - y\|.
\end{align}

Note that, in practice, these terms are intractable, as we would need to compute distances for all $|\mT| = 2^{N \times N \times N}$ possible topologies to estimate the expectancy over $E _{T}$.
One could use Monte Carlo Integration \cite{Niederreiter92} to approximate these quantities. 
However, in order to get an accurate estimate, this would still require us to compute several times the distance between target point cloud and sampled one, which is per-se a costly operation.  
We address this limitation by approximating the expectancy terms with a single realization for each training iteration and relying on Stochastic Gradient Descent, used for training the network, to obtain reliable estimates.   

We therefore approximate equation \ref{eq:loss} with the following terms:
\begin{align}
\mathbb E \left[ \,d(X(T, \Delta) \mid Y) \right]  \simeq & \sum _{n=1} ^{ N \times N\times N} p(t_n = 1) \min_{y \in Y} \| x_n - y\| \\
\mathbb E \left[ \,d(Y\mid  X(T  , \Delta ))  \right] \simeq&  \sum _{y \in Y} \min_{x \in X(T ^*,\Delta ^*)} \| x - y\|,
\end{align}
where $T^*, \Delta ^*$ denote random realizations. 
Note that, this way, we have to compute distances between point clouds only once per loss term evaluation.

In order to guide the learning of a meaningful surface occupancy tensor, we complement the reconstruction loss with the following \textit{binary cross entropy} loss
\begin{align}
&\mathcal L _{\text{BCE}}(O, T = T(Y)) = \\ & \sum_{n \in {N \times N \times N}} (1-t_n) \log (1- o_n)  - t_n \log (o_n),
\end{align}
where $T(Y)$ is a binary tensor which indicates if voxel $n$ contains at least a target point or not. This term encourages occupancy tensor to match voxelized point cloud. Note that, since we randomly sample a target point cloud $Y$ from the ground truth 3D surface at each iteration, the surface occupancy tensor effectively estimates the portion of surface area contained in each cell. That is to say, cells containing large portions of surface area will often contain sample points, whereas cells just barely crossed by the surface will not get points falling within them most of the times. 

\textbf{Consistency loss.} In order to decouple surface generation and sampling, we should enforce latent codes to not include any information about the output 3D coordinate themselves, but rather only contain a description of the geometric surface they encode. 
Observing that the architecture of our encoder helps us naturally achieving this goal, as different input samplings of the same surface will be mapped roughly to the same latent representation thanks to the Grid Pooling operation, we further promote the learning of latent codes that do not contain sampling information by proceeding as follows:
\begin{itemize}
	\item the point cloud  $Y_{\text{out}}$ that our network aims at reconstructing consists of a different surface samplings than the network input one $Y_{\text{in}}$.
	\item we train our network with batches consisting of different surface samplings of the same surface, and enforce latent codes $z$ to be the same across all elements of batch $B$ with the consistency loss term:
	\begin{align}
	\mathcal L _{\text{consistency}} = \sum _ {b \in B} \| z_b - \bar z \| ^2,
	\end{align}
	where $\bar z = \frac{1}{|B|} \sum _ {b \in B} z_b$.
\end{itemize}

% !TEX root = top.tex
% !TEX spellcheck = en-US

\begin{figure*}[t!]
	\begin{center}
		{\includegraphics[width=\textwidth]{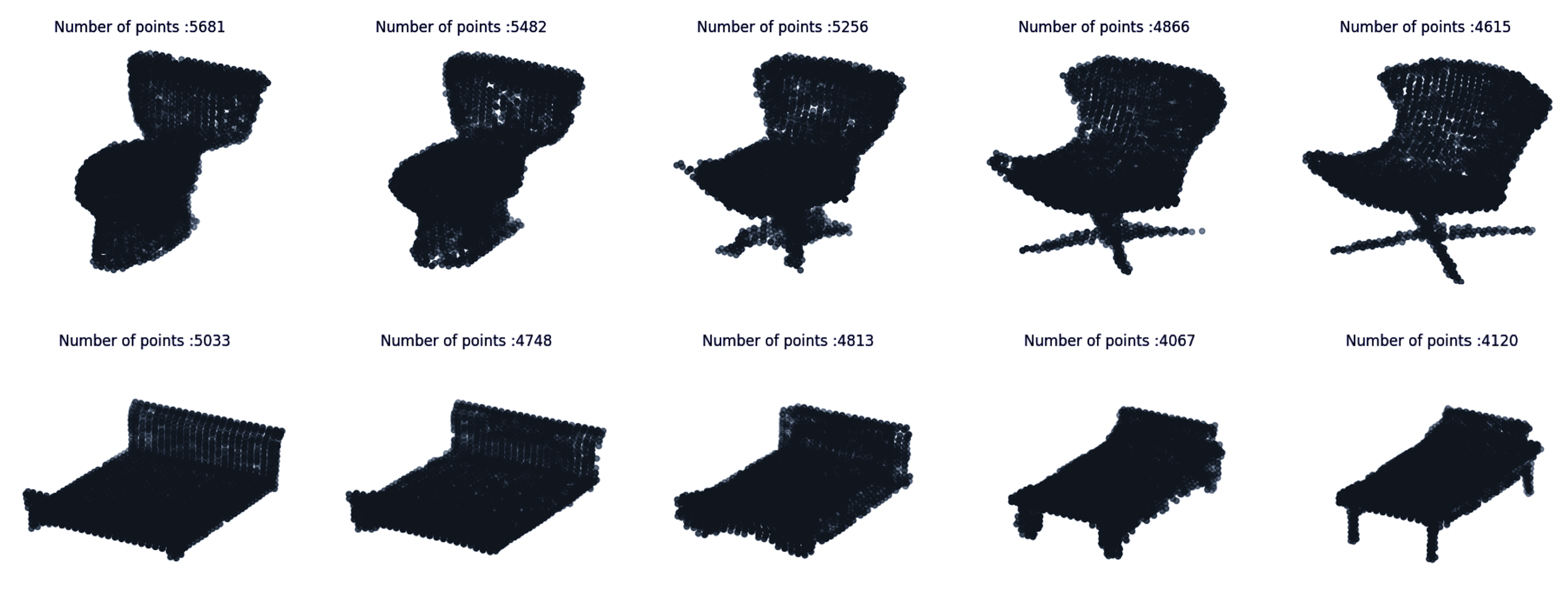}}
		\vskip -0.1in
		\caption{\textbf{Interpolating samples in latent space.} (top) Two shapes within the same class. (bottom) Two shapes from two different classes.  }
		\vskip -0.1in
		\label{fig:interpolation}
	\end{center}
\end{figure*}

\section{Results}
\label{sec:exp}

In this section, we evaluate our technique both qualitatively and quantitatively on standard point cloud representation learning benchmarks. To this end, we use two different tasks that are commonly used to assess algorithms that operate on 3D point clouds: first  latent space interpolation and then clustering and transfer classification. For each one, we report both our results and those of competing methods for which equivalent results have been published. Finally, we 
demonstrate that our latent representation can generate outputs or arbitrarily high resolution and we benchmark our method against state-of-the-art point cloud super resolution algorithms.

\subsection{Training Specifications, Benchmark Datasets, and Metrics}
\label{sec:dataset}

We pre-train our models for a total of 50 epochs on the ShapeNet dataset~\cite{Wu15b}, which consists of 57447 models from 55 categories of man-made objects. For each training iteration we sample 2048 surface points and augment our data by applying random rotations around the gravity axis.  We train our model for a total of 50 epochs, with an initial learning rate of $0.001$ and batch size equal to 43.

To demonstrate the ability of NeuralSampler to learn representations and perform feature extraction, we use the same version of the ModelNet dataset as in~\cite{Yang18a, Achlioptas18}. It comprise 3991/909 models for training/testing for the so-called MN10 split containing 10 classes of man-made objects and 9843/2468 models  for the so-called MN40 split containing 40 classes of man-made objects. 

To showcase NeuralSampler's super-sampling abilities, we use the dataset introduced in~\cite{Yu18a}.  It comprise 60 different models from the VISIONAIR shape repository~\cite{Attene13}, separated into a 40/20 train-test split. They range from smooth 3D models (e.g. animals) to sharp rigid objects (e.g. chairs) . Along with the dataset, the authors proposed two metrics to measure the closeness of 3D point cloud to ground truth meshes. The first one, which we will refer to as \textit{distance} simply is the mean and standard deviation of the distance between output points and the meshes. The second one, the so-called \textit{normalized uniform coefficient} (NUC), quantifies the uniformity of the output point set by measuring the standard deviation in the number of points falling within equal sized disks randomly chosen on the object surface. In practice, given D---9000 in our experiments---equally sized disks, ones first measures the number of points $n_d$ falling within the $d$-th disk, and then define the metric as 
\begin{align}
\text{avg} &= \frac{1}{K D}\sum_{k=1}^K \sum _{d=1}^D  \frac{n_d ^k}{p N^k} ,\\
\text{NUC} &= \left( \frac{1}{K D}\sum_{k=1}^K \sum _{d=1}^D ( \frac{n_d ^k}{p N^k} - \text{avg})^2 \right) ^{1/2},
\end{align}
where $N^k$ is the total number of output points on the $k$-th object, $K$ is the total number of objects, and $p$ is the percentage of disk area over the total object surface area.

\subsection{Latent Space Interpolation}

A generally accepted way to show that a latent representation is valid is to interpolate dataset samples . To create Figure ~\ref{fig:interpolation}, we picked pairs of latent vectors for shapes belonging to the same class and for shapes belonging to different classes. For each pair we computed their weighted averages and decoded the result. In both cases, as we change the relative weights, the decoded shape morphs smoothly to one object to the other. The results demonstrate the smoothness of the learned latent representation.

\subsection{Clustering and Transfer Classification}
\label{sec:classif}

\begin{figure}[h!]
	\begin{center}
		{\includegraphics[width=\columnwidth]{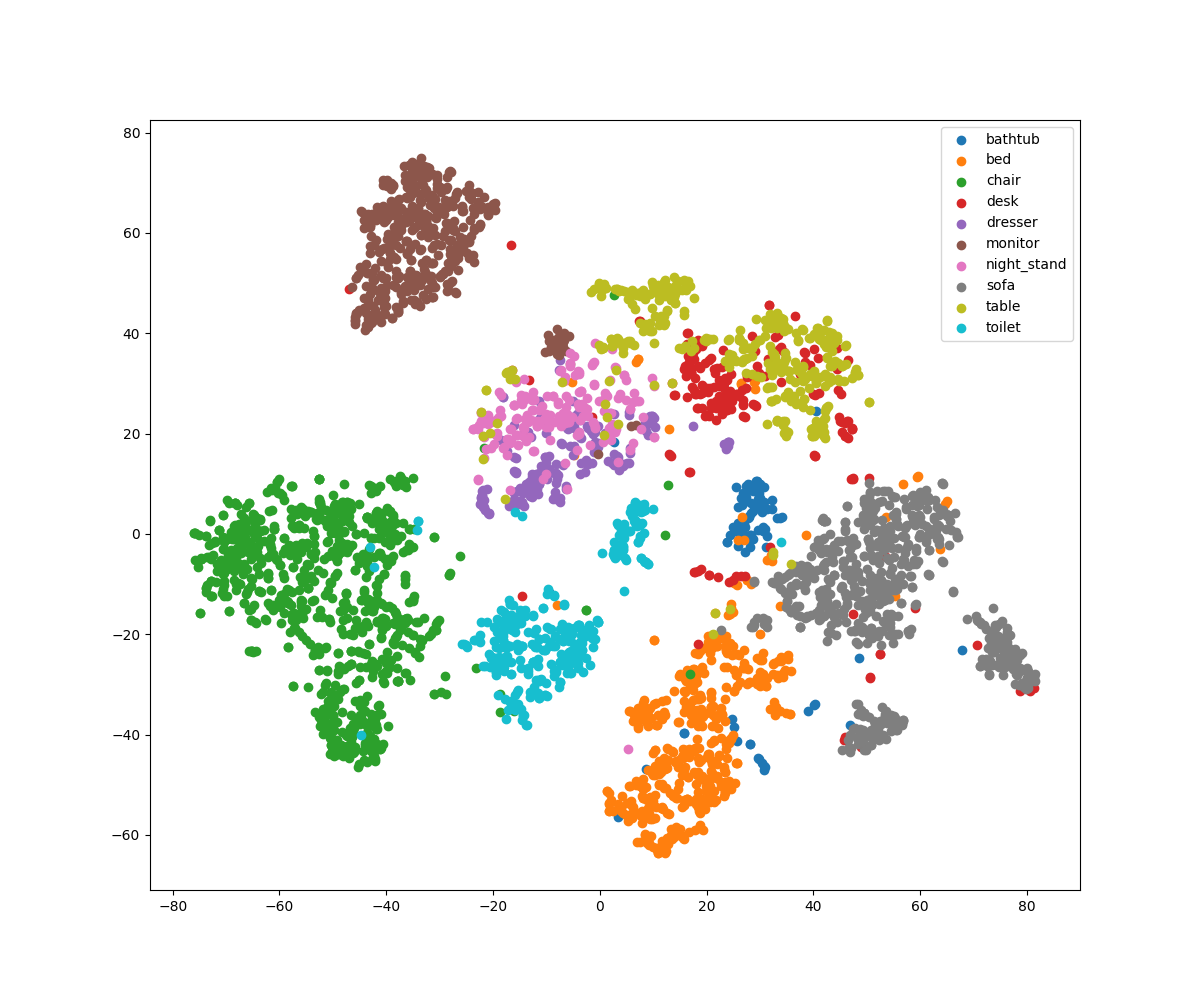}}
		\vskip -0.3in
		\caption{\textbf{T-SNE clusters.} Visualization of latent vectors clustering for the MN10 shapes. }
		\label{fig:clustering}
	\end{center}
	\vskip -0.2in
\end{figure}

Following the same evaluation methodology as in~\cite{Yang18a}, we show in Figure ~\ref{fig:clustering} the clustering obtained for the MN10 dataset by training the NeuralSampler auto-encoder Architecture on the much larger ShapeNet dataset, as discussed above. To create this figure, we used the T-SNE transform \cite{Maaten08} that maps a high dimensional space into a two dimensional one in an iterative fashion. In this case, we set the perplexity parameter to $50$.
Note that most classes are easily separated except, as already  noted in \cite{Yang18a}, the dresser (violet) from the night-stand (pink) and the desk (red) from the table (yellow). Interested readers can convince themselves that these classes are indeed hard to distinguish by visually inspecting the corresponding point clouds.

\begin{table}[h!]
	\begin{tabular}{l | l | l}
		Method  & MN10 & MN40\\
		\hline
		SPH \cite{kazhdan03}          &     79.8 \%     & 68.2 \% \\
		LFD \cite{chen03}              & 79.9  \%    &  75.5 \% \\
		T-L Network \cite{girdhar16}   & -   &   74.4 \% \\
		VConv-DAE  \cite{Sharma18}     & 80.5 \%   &  75.5 \% \\
		3-D GAN  \cite{Wu16b}     &  91.0 \%       &83.3 \% \\
	    Latent GAN \cite{Achlioptas18}  &   \textbf {95.3} \% & 85.7 \% \\
	    Point Cloud GAN \cite{Li18e}  & -   &   87.8 \% \\
		FoldingNet \cite{Yang18a}     &  94.4 \% &  {88.4} \% \\
		NeuralSampler (Ours)           &    \textbf {95.3} \%      & \textbf {88.7} \% \\
	\end{tabular}
	\caption{ Classification accuracy on ModelNet~\cite{Wu16b}. For all methods, we train a linear SVM on the latent vectors these unsupervised learning methods produce.}
	\label{tab:classification}
\end{table}

To demonstrate the power of our auto-encoder in learning meaningful representations, we use the benchmarking pipeline proposed in~\cite{Wu16b, Achlioptas18}. That is, we first pre-train our auto-encoder on the in ShapeNet as discussed in Section~\ref{sec:dataset}. We then assess the quality of the learned embeddings by using the trained encoder to compute latent codes for the ModelNet dataset, and train a linear SVM to classify them into shape categories.

To guarantee a fair comparison with earlier approaches, we adopt the same training/test split as in~\cite{Wu16b} and take our auto-encoder bottleneck to be of dimension 512 as in~\cite{Yang18a,Achlioptas18}.  We report our results on the MN10 and MN40 versions of ModelNet in  Table \ref{tab:classification}. On MN10, our model yields performance comparable to that of~\cite{Achlioptas18} while using twenty times fewer parameters in the decoder structure. On MN40,  dataset our model outperforms all the others and uses half as many parameter for its decoder as the architecture of~\cite{Yang18a} that comes second. To be more precise, the number of parameters in the fully connected decoder of~\cite{Achlioptas18} is $1.52 \times 10^{7}$, while the folding net decoder architecture of ~\cite{Yang18a} has $1.05 \times 10^{6}$ parameters. Our approach exploits an Euclidean grid to reduce the number of parameters to $4.63 \times 10^{5}$.

\subsection{Point Cloud Super-Resolution}

\begin{figure}[h!]
	\begin{center}
		{\includegraphics[width=\columnwidth]{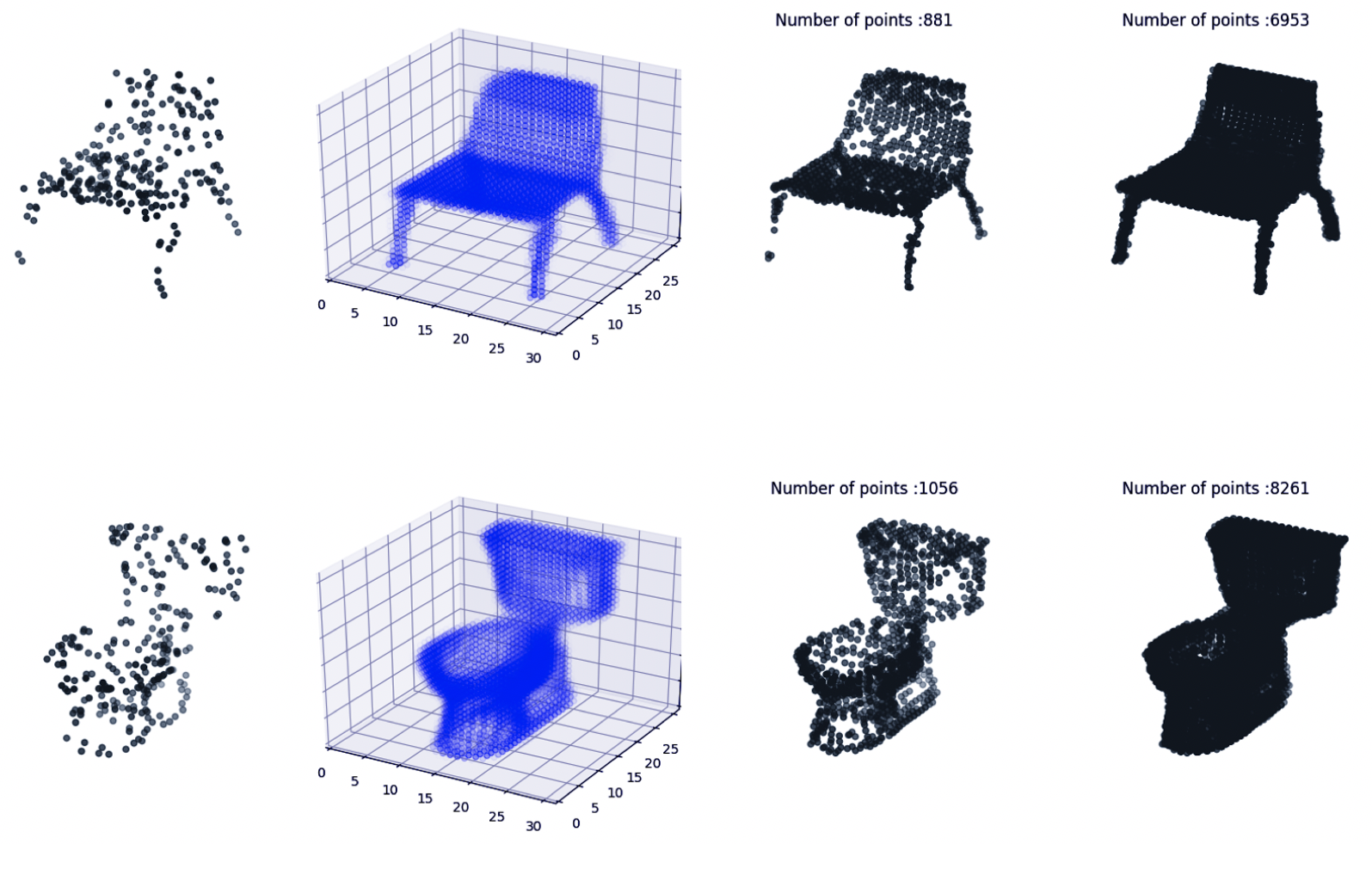}}
		\vspace{-0.8mm}
		\caption{\textbf{Cloud super-sampling.} Our decoder operates on a low resolution grid, which makes our approach directly usable for cloud super-sampling. In this case we trained our model with 2048 input sample points but feed only 256 at test time. We show from the left to right  input point cloud, occupancy map, point clouds generated by 1 and 8 forward passes. Note that unlike in Figure~\ref{fig:sampling} where the input clouds were dense, here they are very sparse without appreciable degradation in the output quality. }
		\label{fig:resolution}
	\end{center}
	\vskip -0.2in
\end{figure}

We now turn to increasing the resolution of a sparse cloud of 3D points as shown in Figure ~\ref{fig:resolution}.

%In this section we demonstrate how our auto-encoder architecture can be naturally utilized for point cloud super resolution applications, thanks to the specific design of both the encoder and the decoder architecture. The grid pooling operation allows our encoder to produce consistent embeddings even when operating with point clouds having a much lower resolution than the ones it was originally trained on, as also noted in \cite{Liao18a}, This favorable property is further enhanced by the fact that in our application we work with a grid that has very low spatial resolution with respect the the point clouds it handles. The Sampling Layer in our decoder then allows us to produce output point clouds with an arbitrary number of elements, as we can just iteratively sample from it until our output has the desired spatial resolution. We can therefore use our auto-encoder effectively for point cloud super-resolution applications, as we illustrate in  without modifying the network architecture or the way we train it.

\begin{table}[ht!]
	\centering
	\begin{tabular}{l | l | l l}
	        \hline
		\multirow{2}{*}{Method}  & \multicolumn{2}{c}{Distance ($10^{-2}$)}  \\
		\cline{2-3}
		& mean & std   \\
		\hline
		PointNet \cite{Qi17a}   & 2.27 & 3.18      \\
		PointNet++ \cite{Qi17b}   & 1.01 & 0.83      \\
		PointNet++(MSG) \cite{Yu18a}  & 0.78 & 0.61  \\
		PUNet  \cite{Yu18a} & 0.63 & 0.51  \\
		NeuralSampler (Ours) & \textbf {0.54}& \textbf {0.41}  \\
		\hline
	\end{tabular}
	\vspace{-2mm}
	\caption{ Super-resolution accuracy  on the dataset of~\cite{Yu18a}.}
	\label{tab:super_dist}
\end{table}

\begin{table*}[t!]
	\begin{center}
		\begin{tabular}{l | l | l || l | l | l | l | l | l}
			\hline
			\multirow{2}{*}{Method} & \multicolumn{6}{c} {Normalized Uniformity Coefficient, p= }  \\
			\cline{2-7}
			& 0.2\%  & 0.4\% & 0.6\%  & 0.8\% & 1.0\%  & 1.2\%\\
			\hline
			PointNet \cite{Qi17a}         &    0.409  & 0.334 & 0.295 & 0.270 & 0.252 & 0.239 \\
			PointNet++ \cite{Qi17b}       &    0.215  & 0.178 & 0.160 & 0.150 & 0.143 & 0.139  \\
			PointNet++(MSG) \cite{Yu18a}      &    0.208  & 0.169 & 0.152 & 0.143 & 0.137 & 0.134\\
			PUNet  \cite{Yu18a}     &   \textbf {0.174}  & \textbf {0.138} & 0.122 & 0.115 & \textbf {0.112} & 0.110  \\
			NeuralSampler (Ours)    &    0.222  & 0.144 & \textbf {0.116} & \textbf {0.108} & 0.113 & \textbf {0.100}  \\
			\hline
		\end{tabular}
	\end{center}
	\caption{ Super-resolution uniformity on the dataset of~\cite{Yu18a}.}
	\label{tab:super}
\end{table*}

We evaluate quantitatively our approach by using the distance and NUC metrics, as defined in Section~\ref{sec:dataset}, on the dataset proposed in~\cite{Yu18a}. As in Section~\ref{sec:classif}, we pre-train on ShapeNet and refine the VISIONAIR training set by running 100 epochs and applying random rotations to the input point clouds.  At test time, we feed input point sets with $1024$ elements, and, as in~\cite{Yu18a}, output point sets with roughly $4000$ elements by performing 4 forward passes.

We report our results in Table~\ref{tab:super_dist}, which shows that once again we outperform the state-of-the-art in terms of accuracy. In Table~\ref{tab:super}, we report the corresponding NUC metric. Even though our method is more accurate than that of~\cite{Yu18a}, the second-best in terms of this test, it is very similar in terms of sampling regularity, which is what NUC measures. We attribute this to the fact that~\cite{Yu18a}  includes a repulsion loss term between points during training that  explicitly favors uniformity whereas we do not. Our approach nevertheless achieves sampling uniformity because it relies on an underlying grid structure.

% !TEX root = top.tex
% !TEX spellcheck = en-US

\section{Conclusion}
\label{sec:conclusions}

We have proposed an auto-encoder architecture that can both encode and decode clouds of arbitrary size and demonstrated its effectiveness at upsampling sparse point clouds. Given a 3D point cloud that represents a surface and a corresponding bounding cube, it computes a latent vector representation that encodes, for each voxel of the bounding cube, both the area of the surface patch that falls within the voxel and the position of the patch with respect the voxel center.  

This latent vector representation is key to being able to properly upsample the surface. In future work, we will use it to build generative models and model surfaces of arbitrary topology that require upsampling to accurately represent arbitrarily complex objects. This should prove valuable to address shape-from-X problems, such as the reconstruction of deforming clothes from video sequences. For such potentially complex deformations, it is often difficult to guess {\it a priori} where the model's resolution should be augmented, for example because wrinkles or folds are appearing. In such a scenario, the ability to dynamically change the resolution should prove invaluable.

\bibliography{string,vision,graphics,learning,misc}
\bibliographystyle{icml2019}

%%%%%%%%%%%%%%%%%%%%%%%%%%%%%%%%%%%%%%%%%%%%%%%%%%%%%%%%%%%%%%%%%%%%%%%%%%%%%%%
%%%%%%%%%%%%%%%%%%%%%%%%%%%%%%%%%%%%%%%%%%%%%%%%%%%%%%%%%%%%%%%%%%%%%%%%%%%%%%%

\end{document}